\documentclass[runningheads]{llncs}
\usepackage{graphicx}
\usepackage{xcolor}
\usepackage{color, soul}
\usepackage{arydshln}
\begin{document}
\title{Synthetic and Real Inputs for Tool Segmentation in Robotic Surgery}
%
%
\author{Emanuele Colleoni\inst{1, 2}\orcidID{0000-0003-4614-5742} 
\and
Philip Edwards\inst{1, 2}\orcidID{0000-0003-0203-5736} 
\and
Danail Stoyanov\inst{1, 2}\orcidID{0000-0002-0980-3227}}
%
\authorrunning{E. Colleoni et al.}
%
\institute{Wellcome/EPSRC Centre for Interventional and Surgical Sciences (WEISS), University College London, 43-45 Foley St., Fitzrovia, London W1W 7EJ, UK \and
Department of Computer Science, University College London, United Kingdom
\email{emanuele.colleoni.19@ucl.ac.uk}\\
}
\maketitle              
\begin{abstract}
Semantic tool segmentation in surgical videos is important for surgical scene understanding and computer-assisted interventions as well as for the development of robotic automation. The problem is challenging because different illumination conditions, bleeding, smoke and occlusions can reduce algorithm robustness. At present labelled data for training deep learning models is still lacking for semantic surgical instrument segmentation and in this paper we show that it may be possible to use robot kinematic data coupled with laparoscopic images to alleviate the labelling problem. We propose a new deep learning based model for parallel processing of both laparoscopic and simulation images for robust segmentation of surgical tools. Due to the lack of laparoscopic frames annotated with both segmentation ground truth and kinematic information a new custom dataset was generated using the da Vinci Research Kit (dVRK) and is made available.

\end{abstract}
\section{Introduction}

Robotic minimally invasive surgery is now an established surgical paradigm across different surgical specialties~\cite{palep2009robotic}. While the mechanical design and implementation of surgical robots can support automation and advanced features for surgical navigation and imaging, significant effort is still needed to automatically understand and infer information from the surgical site for computer assistance. Semantic segmentation of the surgical video into regions showing instruments and tissue is a fundamental building block for such understanding~\cite{moccia2018uncertainty,kurmann2017simultaneous} and to pose estimation for robotic control~\cite{allan20183,du2018articulated} and surgical action recognition~\cite{tao2013surgical}.

The most effective semantic segmentation approaches for surgical instruments have used deep learning and Fully Convolutional Neural Networks~(FCNNs)~\cite{garcia2017toolnet}. Various architectures have been reported including novel encoder-decoders using established pre-trained feature extractors or adding attention fusion modules in the decoding part of the network~\cite{laina2017concurrent,ni2019rasnet}. These have dramatically improved algorithm performance compared to early methods~\cite{bouget2017vision}. More recently, robotic systems articles have also reported the coupling of visual information with kinematic data~\cite{su2018real} and the possibility of using kinematic information to produce surgical tools segmentation ground truth~\cite{da2019self}. In their work, da Costa Rocha et al.~\cite{da2019self} employed a Grabcut-based cost function to iteratively estimate the optimal pose of the kinematic model in order to produce accurate segmentation labels through tool's model projection on the image plane. The major weakness of this method is its strong dependence to an accurate initial pose of the tool, that is not trivial in the surgical scenario. A similar approach was attempted by Qin et al.~\cite{qin2019surgical}: their method is based on a particle filter optimization that repeatedly updates the pose of the tool to match the silhouette projection of the surgical tool with a vision-based segmentation mask obtained using a ToolNet~\cite{garcia2017toolnet}. However, this procedure heavily rely on optical markers and on a navigation system for initial tool pose estimation. Moreover, the procedure has been proposed only for non-articulated rigid tools, that limits the field of applicability of this method.  Despite progress, robust semantic segmentation for surgical scene understanding remains a challenging problem with insufficient high quality data to train deep architectures and more effort needed to exploit all the available information on instrument geometry or from robotic encoders.

In this paper, we propose a novel multi-modal FCNN architecture that exploits visual, geometric and kinematic information for robust surgical instrument detection and semantic segmentation. Our model receives two input images: one image frame recorded with a da Vinci® (Intuitive Surgical Inc, CA) system and a second image obtained loading the associated kinematic data into a virtual da Vinci Research Kit~(dVRK) simulator.
The global input is an image couple showing real (containing visual features) and simulated (containing geometric tools information from robot Computer-Aided Design~(CAD) models) surgical tools that share the same kinematic values. We show that the simulation images obtained exploiting kinematic data can be processed in parallel with their real counterpart to improve segmentation results in presence of variable light conditions or blood. This is the first attempt that uses a deep learning framework for parallel processing of images produced using a robot simulator and a laparoscopic camera to improve surgical tool segmentation avoiding iterative shape matching. Due to a lack of a dataset annotated with both kinematic data and segmentation labels, we built a custom dataset of 14 videos for the purpose\footnote{https://www.ucl.ac.uk/interventional-surgical-sciences/davinci-segmentation-kinematic}.

\section{Methods}\label{sec: methods}


\begin{figure}
\includegraphics[width=\textwidth]{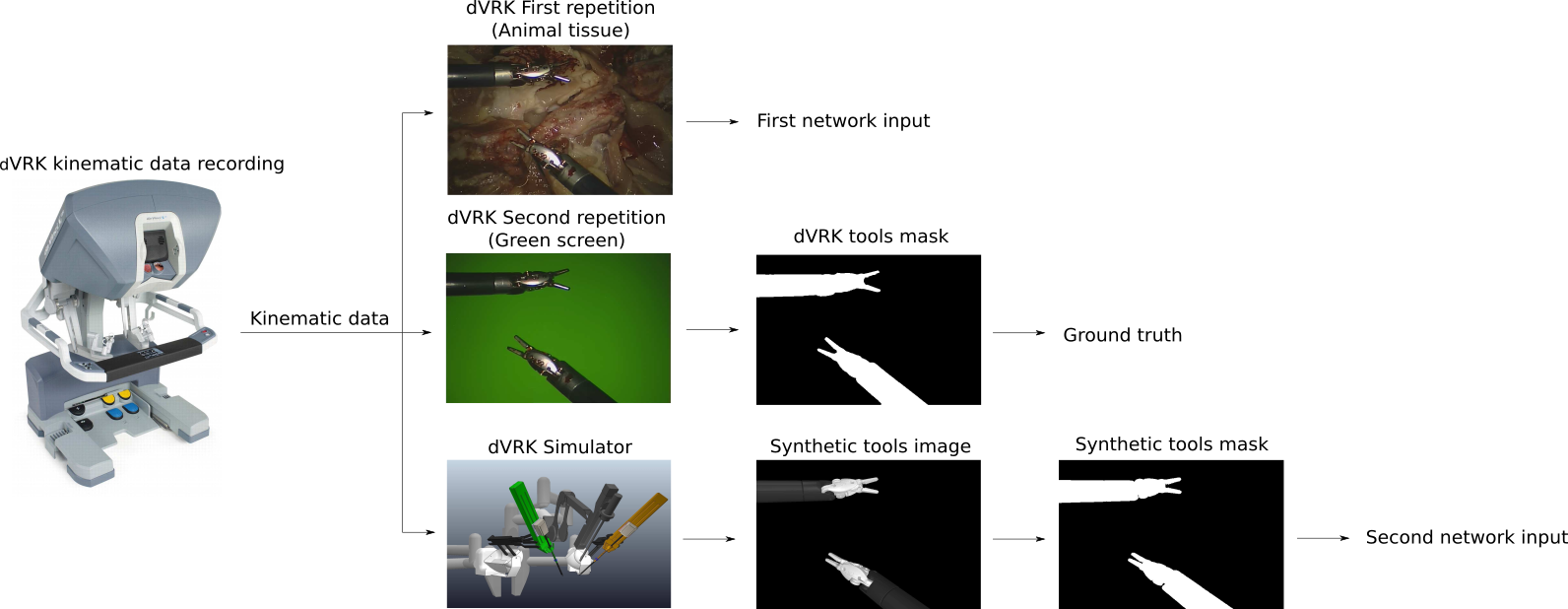}
\caption{The figure shows the workflow for the generation of our dataset. Once the kinematic data of a movement is recorded using the dVRK, it is first reproduced over an animal tissue background. A second repetition with the same kinematic is then performed on an OLED green screen. The ground truth for each image is the generated using background subtraction technique.
The collected kinematic data are then loaded on a dVRK simulator to produce simulation images of the tools, that are successively binarized to produce the second  input of the proposed FCNN.}\label{fig: ground_truth}
\end{figure}

\subsection{Dataset Generation with dVRK}\label{sec: dataset_gen}

We use the dVRK system to record both video and kinematic information about the instrument motion. Because the system is robotics we can repeat movements previously executed by an operator on a da Vinci Surgical System\footnote{https://www.intuitive.com/} (DVSS) in clinical practice. Each instrument on the dVRK held by the Patient Side Manipulators (PSM) is defined by 7 joints (6 revolute and 1 prismatic), while the Endoscope Control Manipulator has 4 (3 revolute and 1 prismatic). In this study we only use EndoWrist® Large Needle Drivers for simplicity, although the same workflow can be extended to the whole family of articulated surgical tools if appropriate models and control information is available (currently not implemented in dVRK).

To produce each video in the dataset we followed four consecutive steps:
\begin{itemize}
  \item First, we perform a surgical movement recording kinematic data using our dVRK;
  \item Then we collect image frames with animal tissue background using the recorded kinematic data stream;
  \item The same movement is reproduced a second time on a green screen to obtain tools ground truth segmentation masks;
  \item Finally, for each frame, we produced an associated image of the virtual tools obtained employing a dVRK simulator by making use of the recorded kinematic information.
\end{itemize}
Our dataset generation procedure is shown in Fig.~\ref{fig: ground_truth}.

\subsubsection{Kinematic Data}\label{subsec: kin_rec}
An action is first performed on the DVSS without a background and the kinematic information of the PSMs and ECM is recorded. The recording framework was implemented in MATLAB using the Robotic System Toolbox\footnote{https://uk.mathworks.com/products/robotics.html} to access robot articulations joint values from the dVRK. The result is a 7~by~N matrix for each PSM and 4~by~N for the ECM, where each row corresponds to one joint (starting from the ECM/PSM's base till the tip of the tool) and consecutive columns represent consecutive time steps.
\subsubsection{Video Data}
The video frames are collected by repeating the recorded action over an animal tissue background. The stored joint coordinates of all PSMs and ECM are sent to the DVSS via the dVRK using our MATLAB interface in order to have precise movement reproduction. Video images are synchronously collected every 150 ms to avoid redundancy in the data.

\subsubsection{Ground Truth Generation}
The segmentation ground truth is produced by physically replacing the animal tissue background with a green screen. We chose to use an Organic Light-Emitting Diode~(OLED) screen emitting green light to avoid shadows that generally decrease segmentation performances. Once the screen is conveniently placed to entirely cover the camera Field Of View~(FOV), the same recorded action employed in the previous phase is reproduced. Finally we removed the tools from the FOV and an image is collected showing only the background. The segmentation ground truth for each frame is then obtained by subtracting the background image to each frame and by applying a threshold to the \textit{L1-norm} of the subtraction result.

The ground truth generation procedure allows us to avoid issues originated by a virtual replacement of the segmentation mask background, such as image matting and blending~\cite{chuang2001bayesian}. The reliability of our ground truth generation methodology has been tested on 7 further video couples, where a same action was reproduced twice on the green background. The resulting ground truth masks for each video couple were then compared using Intersection over Union~(IoU) metric, obtaining an overall 99.8\% median evaluation score with Interquartile Range~(IQR) of 0.05\%.

\subsubsection{Simulation Images} We load the kinematic data collected using our dVRK into the simulation model ~\cite{fontanelli2018v} to virtually reproduce the performed movement in CoppeliaSim\footnote{https://www.coppeliarobotics.com/}\cite{rohmer2013v}. Images were simultaneously collected at the same frame rate used for the recorded videos in order to have synchronization between simulation and dVRK information. The produced images were then thresholded to obtain a segmentation mask used to feed the proposed network.

\begin{figure}
\includegraphics[width=\textwidth]{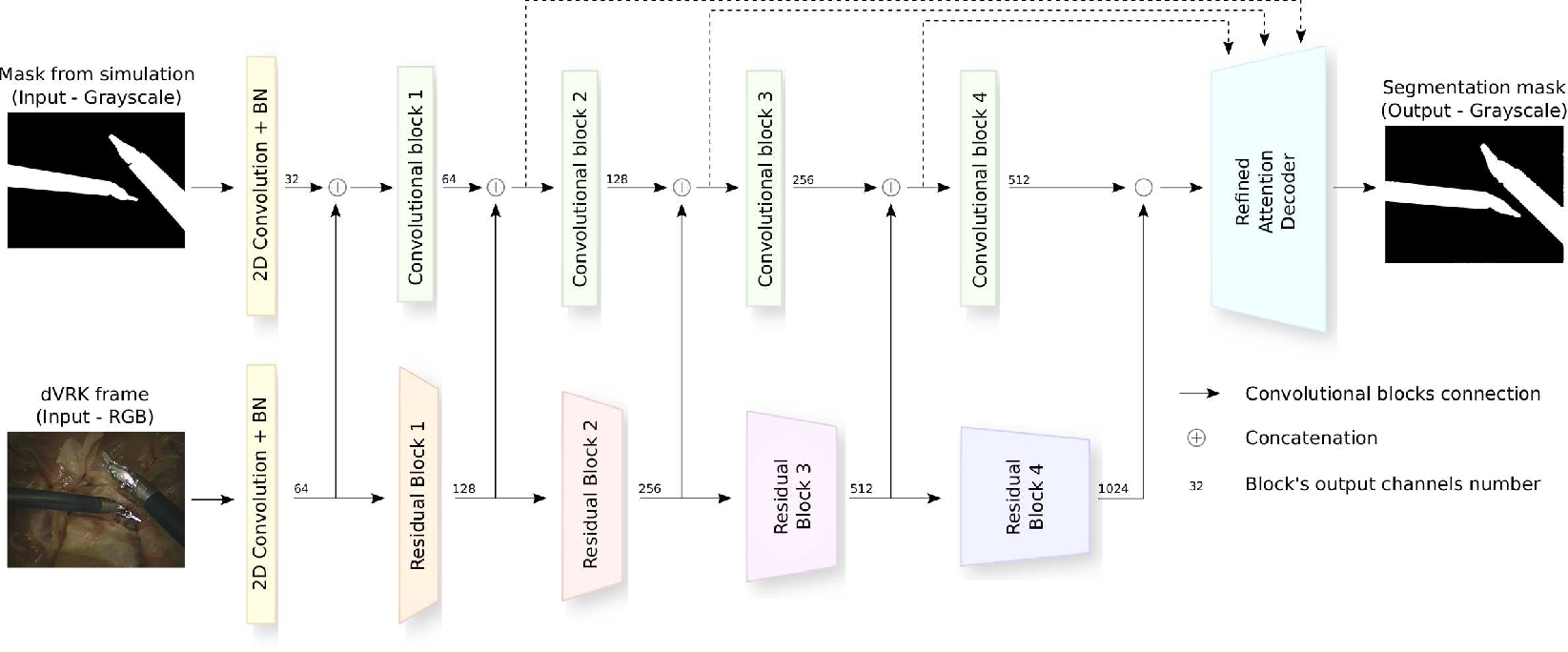}
\caption{Architecture of the proposed model. Two connected branches are used to extract features from dVRK frames (RGB) and simulated tools segmentation masks (Grayscale). The lower branch is made of residual blocks developed in~\cite{he2016deep} pre-trained on Imagenet dataset, while the upper branch is composed of 4 convolutional blocks, each one performing 3 convolutions + batch normalization + RELU activation. The number of output channels for all encoders' blocks is written at the end of  each one of them. The decoder part is described in~\cite{ni2019rasnet}, taking encoder's output and skip connections as input.}\label{fig: architecture}
\end{figure} 

\subsection{Network Architecture}\label{sec: net_arch}
We propose a double-input FCNN for simultaneous processing of frames collected using the dVRK and segmentation masks of their simulated counterpart. The architecture of the proposed network is shown in Fig.~\ref{fig: architecture}. We a used the commonly adopted U-net structure as starting point for our model ~\cite{ronneberger2015u}. We chose to concatenate features extracted from both inputs from a very early stage in the network, that has shown to be more effective than merging them only in the decoding part ~\cite{guo2019deep}.

The encoder branch for dVRK frames processing was implemented following work conducted in~\cite{ni2019rasnet} and~\cite{iglovikov2018ternausnet}, where features are extracted from the image using ResNet50~\cite{he2016deep} blocks pre-trained on Imagenet dataset~\cite{deng2009imagenet}. Each residual block consists of multiple consecutive sub-blocks, namely 3, 4, 6 and 3 sub-blocks for residual block 1 to 4 respectively. Each sub-block is composed by 3 convolutional layers, where a Batch Normalization~(BN) and a Rectified Linear Unit~(RELU) activation are applied on the output of each convolution stage. The resulting output is concatenated with the input of the sub-block using a skip connection. 
We built a second branch parallel to ResNet50 for simulated tools' mask processing. Following ResNet50, a 7x7 convolution + BN + RELU activation operation is first performed on the image. The result is then passed through 4 consecutive convolutional blocks. At each block, the input coming from the previous convolutional block is first concatenated with the output of the relative parallel residual block. The result is then processed using 3 different convolutional layers, i.e. a first 1x1 convolution with stride 1 to double the number of channels, followed by a 3x3 convolution (stride 1x1) and a final 3x3 convolution with stride 2x2 to halve the output's height and width.

The segmentation probability map is finally obtained concatenating the outputs of both the last convolutional and residual blocks and processing them employing the decoding architecture (based on attention fusion modules and decoder blocks)developed in~\cite{ni2019rasnet}. Each attention block takes as second input the features extracted from intermediate encoding layers of both branches using skip connection. Such methodology has shown to be particularly useful to properly recover information lost in the network's early blocks~\cite{ronneberger2015u}.

\subsection{Loss Function Details}\label{sec: loss_met}

We selected the sum of per-pixel binary~crossentropy~\cite{ni2019rasnet,iglovikov2018ternausnet} which has been previously used for instrument detection and articulation tracking ~\cite{du2018articulated,mohammed2019streoscennet}, and IoU loss (to prevent the network to be sensitive to class unbalance) as loss function to train our model.  The binary ~crossentropy is defined as:
\begin{equation}
 \textit{$L_{bce}$} = \frac{1}{\Omega}\sum_{n\in\Omega}^{} [p_nlog\widehat{p}_n + (1-p_n)log(1-\widehat{p}_n)]
\end{equation}

where $p_n$ and $\widehat{p}_n$ are the FCNN output and ground truth values of pixel \textit{n} into mask domain $\Omega$.

\textit{IoU} loss is defined as:

\begin{equation}
  \textit{$L_{IoU}$} = 1 - \textit{${IoU}_{score}$}
\end{equation}

and \textit{$IoU_{score}$} is defined as:
\begin{equation}
  \textit{$IoU_{score}$} = \frac{TP}{TP + FP + FN}
\end{equation}

where TP is the number of pixels correctly classified as tools' pixels, while FP and FN are the numbers of pixels mis-classified as tools and background respectively. Following~\cite{iglovikov2018ternausnet}, we chose a threshold value of 0.3 to binarize our output probability mask.

The resulting loss function is then defined by the sum of \textit{$L_{bce}$} and \textit{$L_{Iou}$}:
\begin{equation}
  \textit{L} = \textit{$L_{bce}$} + \textit{$L_{IoU}$}
\end{equation}

\section{Experiments and Results}

\subsubsection{Dataset}\label{subsec: exp_dataset} We collected 14 videos of 300 frames each (frame size = 720x576), for a total amount of 4200 annotated frames. In particular, 8 videos were used for the training phase, 2 for validation and the remaining 4 for testing. We employed 5 different kinds of animal tissues (chicken breast and back, lamb and pork loin, beef sirloin) for the entire dataset, changing the topology of the background and varying illumination conditions in each video to increase data variability. Lamb kidneys and blood were placed in the background and on the tools of the test set videos in order to properly test algorithms' performance on conditions not seen in the training phase. Finally, following~\cite{du2018articulated}, we added Fractional Brownian Motion noise\footnote{https://nullprogram.com/blog/2007/11/20/} to simulate cauterize smoke on test set frames. Each frame has been first cropped to remove dVRK side artefacts and then resized to 256x320 to reduce its processing computational load. We produced a simulation segmentation mask (see Sec.~\ref{sec: methods}) for each frame using the dVRK simulator in order to feed our double-input FCNN. All data will be made available for research.

\begin{table}[t]

\caption{Comparison with the state of the art architectures on our test set~\cite{iglovikov2018ternausnet}~\cite{ni2019rasnet}. Results for each video are presented in terms of median Intersection over Union~\textit{IoU} score and Interquartile Range~(IQR) over all video frames in percentages.}\label{tab: results}
\resizebox{\textwidth}{!}{
\begin{tabular}{lcccccc}
\multicolumn{7}{c}{\textbf{Median Value (\%) / IQR (\%) of \textit{IoU} score}}                                                                                                      \\ \hline
                                      & \multicolumn{3}{c}{\textbf{No smoke}}                                   & \multicolumn{3}{c}{\textbf{Added Smoke}}                  \\ \hline
                                      & \textbf{Ternausnet} & \textbf{Rasnet} & \textbf{Proposed}               & \textbf{Ternausnet} & \textbf{Rasnet} & \textbf{Proposed} \\ \hline
\multicolumn{1}{l|}{\textbf{Video 1}} & 73.41/12.24         & 79.25/8.84      & \multicolumn{1}{c}{\textbf{81.80/7.74}} & 24.39/9.29          & 44.85/15.80     & \textbf{54.15/14.16}       \\ \hdashline
\multicolumn{1}{l|}{\textbf{Video 2}} & 73.94/2.88          & 78.44/3.80      & \multicolumn{1}{c}{\textbf{81.79/3.71}} & 58.87/7.15          & 73.98/9.41      & \textbf{79.49 4.31}        \\ \hdashline
\multicolumn{1}{l|}{\textbf{Video 3}} & 78.57/6.13          & 84.56/3.97      & \multicolumn{1}{c}{\textbf{91.04/2.79}} & 78.02/6.28          & 85.62/4.41      & \textbf{91.20/3.41}        \\ \hdashline
\multicolumn{1}{l|}{\textbf{Video 4}} & 92.07/3.82          & 95.09/2.15      & \multicolumn{1}{c}{\textbf{95.16/2.68}} & 49.65/10.84         & 78.06/17.28     & \textbf{82.93/8.62}       
\end{tabular}}
\end{table}

\subsubsection{Implementation and Runtime Analysis}\label{subsec: exp_runtime}The proposed model was implemented in Tensorflow/Keras\footnote{https://www.tensorflow.org/guide/keras} and trained on GPU NVIDIA®~Tesla®~V100. We chose Adam as optimizer for our network~\cite{kingma2014adam}, with a learning rate of 0.001 and exponential decay rates \textit{$\beta$1} and \textit{$\beta$2} of 0.9 and 0.999 respectively. We selected the best model weights considering \textit{IoU} score as evaluation metric obtained on the validation set.

\subsubsection{Comparison Experiments}\label{subsec: exp_comp}We examined the benefit introduced by adding geometric and kinematic information in the network input by comparing our results with the ones obtained using Ternausnet~\cite{iglovikov2018ternausnet} and Rasnet~\cite{ni2019rasnet} architectures after being trained on the proposed training set. Performances were first evaluated on the test set, repeating the prediction a second time on the same frames after adding simulated smoke. We selected \textit{IoU} as the evaluation metric.
As shown in Table~\ref{tab: results}, the proposed model achieved good results on all videos compared to the state of the art, with an overall median \textit{IoU} score of 88.49\% (IQR = 11.22\%) on the test set and a score of 80.46\% (IQR = 21.27\%) when simulated smoke is superimposed to dVRK frames. Rasnet and Ternausnet obtained \textit{IoU} scores of 82.27\%~(IQR~=~11.55\%) and 76.67\%~(IQR~=~13.63\%) respectively on raw test videos, while their median performances decreased to 75.78\%~(IQR~=~23.48\%) and 54.70\%~(IQR~=~34.57\%) with added smoke. Focusing on single videos, the lowest scores were obtained on Video~1 and Video~2 by all the considered architectures, both with or without smoke. The best performances were instead achieved by the proposed model on Video~4 (without smoke) and Video~3, with more than 95\%  and 91\% median \textit{IoU} score respectively.

We carried out a further analysis only on test frames that present tools occlusion to investigate models performances under this particular condition. Results were evaluated using the same evaluation metric employed during previous experiments.
Even in this situation, the proposed model outperformed the other architectures, achieving a median \textit{IoU} of 93.70\%~(IQR~=~1.94\%), superior to both Rasnet (median \textit{IoU}~=~87.45\%, IQR~=~2.80\%) and Ternausnet (median \textit{IoU}~=~84.38\%, IQR~=~5.91\%)

\begin{figure}[t]
\includegraphics[width=\textwidth]{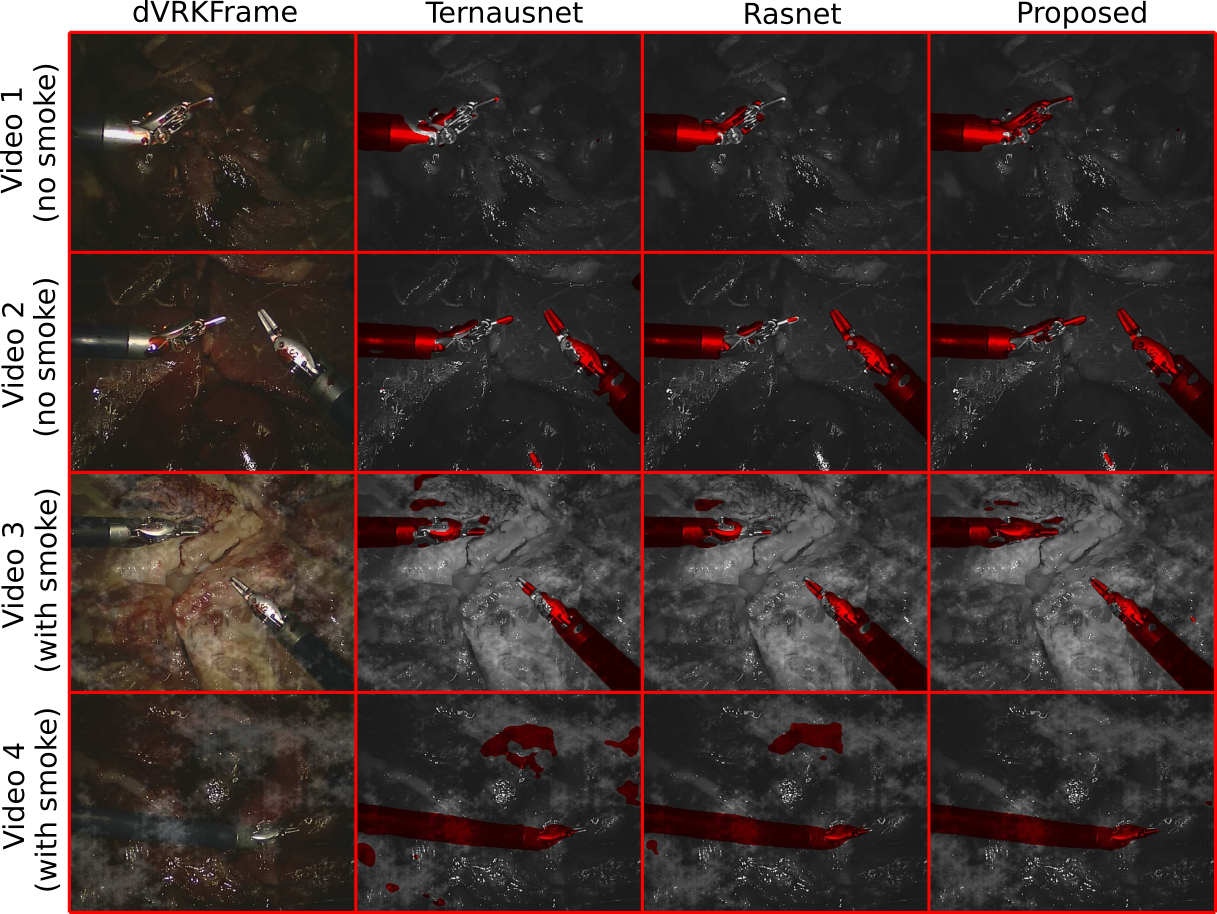}
\caption{Visual example of the obtained results. An image frame for each video (first column) is reported without added smoke (Video 1 and Video 2) and with smoke (Video 3 and Video 4). Segmentation results for Ternausnet, Rasnet and the proposed model are displayed in columns 2, 3 and 4 respectively.}\label{fig: results}
\end{figure}

\subsubsection{Robustness Discussion}\label{subsec: exp_robust}Since no blood on the tools was seen during the training phase, all the architectures learned to label red pixels as background, leading to a mis-classification of the tools' portions covered by blood as seen in Fig.~\ref{fig: results}. Information extracted from segmentation masks of simulated tools helped the proposed network to better recognize the non-covered areas on the tools. Such result is highlighted when smoke is added in the image, with $\Delta$\textit{IoU} scores of 5.51\% and 20.62\% on Video 2 and 5.58\% and 13,18\% on Video 3, compared to Rasnet and Ternausnet respectively. Smoke also deteriorated segmentation performance on videos with poor initial illumination conditions but our model showed good results w.r.t Rasnet and Ternausnet, in this setting. A clear example is shown in Table~\ref{tab: results} for Video 2, where the decrease in performances due to presence of smoke for the proposed model is only 2.3\%, against 4.46\% and 15.07\% suffered from Rasnet and Ternausnet respectively.
The worst results in presence of smoke were obtained on Video~1, where the presence of blood, darkness and smoke led all the considered models to segmentation performances below 55\%~(\textit{IoU} score).
Finally, the behaviour of all the architectures on frames with tool occlusions resembled the one in previous experiments, showing that such scenario do not particularly affect the results.

\section{Discussion and Conclusions}
In this paper, we proposed a double-input FCNN for segmentation of surgical tools from laparoscopic images. Our model takes as inputs both dVRK frames and segmentation masks produced using a dVRK simulator. Each mask is generated by projecting the simulated tools, conveniently positioned using dVRK kinematic data, onto the image plane. Our method achieved state of the art performance against image only-based models, suggesting that geometric and kinematic data can be employed by deep learning frameworks to improve segmentation. We produced a new dataset with segmentation labels and kinematic data for the purpose. Unfortunately, our procedure allows us to produce only binary ground truth. However, it could be interesting to improve the methodology to generate semantic labels in the future.

At the best of our knowledge, this is the first attempt to join visual and kinematic features for tool segmentation using a multi-modal FCNN, avoiding iterative shape-matching algorithms~\cite{da2019self,qin2019surgical}. A further comparison with these methods could however be taken in consideration as future work.
Several ways could be investigated as well to improve our method, e.g. using a residual-learning modelling approach to estimate the difference between simulated and estimated tool masks or trying different architectures~\cite{guo2019deep} to better exploit segmentation input.
Moreover, it could be interesting to study the performances of the proposed model when noise is added to the kinematic data.
Finally, a scenario with tool-tissue interaction could be of great interest from the dataset generation point of view as well for further evaluation analysis in such conditions.

\subsubsection{Acknowledgements}

The work was supported by the Wellcome/EPSRC Centre for Interventional and Surgical Sciences (WEISS) [203145Z/16/Z]; Engineering and Physical Sciences Research Council (EPSRC) [EP/P027938/1, EP/R004080 /1, EP/P012841/1]; The Royal Academy of Engineering Chair in Emerging Technologies Scheme; and Horizon 2020 FET (GA 863146). We thank Intuitive Surgical Inc and the dVRK community for their support of this work.


%
%

\bibliographystyle{splncs04}
\bibliography{Manuscript.bib}
\end{document}